\documentclass{article}

\PassOptionsToPackage{numbers, compress}{natbib}

\usepackage{epsfig, fixmath, amsmath, amssymb, dsfont, xcolor, bm, makecell, floatrow, blindtext, booktabs, subcaption, caption}

\usepackage[final]{neurips_2021_workshop}

\usepackage[utf8]{inputenc} 
\usepackage[T1]{fontenc}    
\usepackage{hyperref}       
\usepackage{url}            
\usepackage{booktabs}       
\usepackage{amsfonts}       
\usepackage{nicefrac}       
\usepackage{microtype}      
\usepackage{wrapfig}
\usepackage{xcolor}         
\newcommand\sr[1]{\textcolor{black}{#1}} 
\newfloatcommand{capbtabbox}{table}[][\FBwidth]

\title{Distribution Estimation to Automate Transformation Policies for Self-Supervision}

\author{%
  Seunghan Yang\thanks{These two authors contributed equally.} , Debasmit Das\footnotemark[1] , Simyung Chang, Sungrack Yun, Fatih Porikli\\
  Qualcomm AI Research\thanks{Qualcomm AI Research is an initiative of Qualcomm Technologies, Inc.}\\
  \texttt{\{seunghan, debadas, simychan, sungrack, fporikli\}@qti.qualcomm.com} \\
}

\begin{document}

\maketitle

\begin{abstract}

In recent visual self-supervision works, an imitated classification objective, called pretext task, is established by assigning labels to transformed or augmented input images. The goal of pretext can be predicting what transformations are applied to the image. However, it is observed that image transformations already present in the dataset might be less effective in learning such self-supervised representations. Building on this observation, we propose a framework based on generative adversarial network to automatically find the transformations which are not present in the input dataset and thus effective for the self-supervised learning. This automated policy allows to estimate the transformation distribution of a dataset and also construct its complementary distribution from which training pairs are sampled for the pretext task. We evaluated our framework using several visual recognition datasets to show the efficacy of our automated transformation policy.

\end{abstract}

\section{Introduction}

Recently, self-supervised learning (SSL) has received great attention in the field of computer vision. In contrast to the supervised learning that requires ground-truth labels, SSL learns representations by defining a pretext task that helps construct labels from the input signals themselves. In literature, a number of pretext tasks have been proposed with various transformations and augmentations: {\it e.g.}, predicting the rotation degrees~\cite{feng2019self,gidaris2018unsupervised}, solving jigsaw puzzles~\cite{noroozi2016unsupervised}, and minimizing the distance between representations of different augmented views of the same image (instance discrimination)~\cite{chen2020simple,he2020momentum}. 
The constructed objective functions of the pretext task are applied to learn the representations, which are then re-used for downstream applications \cite{singh2018self, taha2018two, caron2018deep, gidaris2019boosting, rebuffi2020semi, liu2019exploiting,borse2021inverseform}.

The performance of the self-supervised representation on downstream applications depends highly on the choice of the transformations used in the pretext task.
However, there have been only few works that examined the choice of the transformations \cite{pal2019towards, patrick2020multi, tian2020makes}.
Especially, \cite{pal2019towards} expressed a Visual Transformation for Self-Supervision (VTSS) hypothesis that postulates the learned representations would be less useful if the pretext task involves the transformations already present in the dataset. It provided the empirical verification by manually inspecting the dataset, finding the present transformations (e.g., rotation, translation, and scale), and evaluating the effectiveness of the found transformations with multiple downstream classification problems. 
However, even if the VTSS hypothesis is supported, the following two questions still need to be answered to utilize it in deep learning scenarios, "How to determine the transformation already present in the dataset?", " How to identify transformations that are effective for a self-supervision task?"
This paper proposes a novel method to automate the transformation policy for self-supervision tasks to address the above questions. First, we introduce a learning framework to estimate visual transformations.
We learn a parameterized visual transformation function to estimate the distribution of the transformation present in the dataset.
Then, we construct another distribution which is complementary to the estimated one to obtain the transformation instance not included in the dataset.
For instance, if the estimated transformation is known to follow a uniform distribution between certain bounds, such as 0-90 degrees rotation, we can easily determine its complementary distribution ranging from 90-360. However, modeling the transformation distribution of a dataset without such a prior is not straightforward. To handle this challenge, we define a mapping network that estimates the transformation distribution of the dataset with a known distribution (e.g., Gaussian), and the mapping network is learned by adversarial learning \cite{goodfellow2014generative}.
Here, the histogram of the output values from the mapping network represents the transformation distribution, and then we can construct its complementary distribution to sample a transformation instance for the pretext task of self-supervised learning.
Through extensive experiments on various datasets, including MNIST~\cite{726791}, Fashion MNIST~\cite{xiao2017/online}, SVHN~\cite{netzer2011reading}, CIFAR-10~\cite{krizhevsky2009learning}, and CIFAR-100~\cite{krizhevsky2009learning}, we provide empirical evidence to confirm the efficacy of our automated visual transformation policy.

In summary, the contributions of this paper can be summarized as follows: (i) A framework based on a generative adversarial network to obtain the distribution of transformations present in a dataset; (ii) 
Construction of the distribution complementary to the estimated distribution in (i);
(iii) Generation of a useful pretext task for self-supervised learning with the transformation instances sampled from the complementary distribution;
(iv) Comprehensive study, analysis and comparison of the representations learned from our transformation instances on multiple visual recognition benchmarks.

\section{Related Work} 
\label{gen_inst}

\textbf{Self-supervised learning.} With the capability of learning a representation given unlabeled data, many research works on self-supervised learning have recently been presented and applied to various applications in computer vision, natural language processing~\cite{devlin2018bert,wu2019self}, and robotics~\cite{pathak2019self,jang2018grasp2vec,ebert2018robustness,nair2017combining,murali2018cassl}. Most previous works consider manually designing a pretext task involving specific visual transformations. 

\textbf{Analysis on the visual transformation and strategy.}
Only a few works address the impact of certain geometric transformations in the pretext task to learn representations for downstream tasks \cite{pal2019towards, patrick2020multi, tian2020makes}. To our knowledge, VTSS~\cite{pal2019towards} is the first to discuss that certain transformations are preferable for defining a pretext task.
We automatically find good transformation parameters for defining pretext tasks and propose an automated policy choosing the transformation instances for the target dataset.

\begin{figure}
\begin{subfigure}{.55\textwidth}
  \includegraphics[scale=0.4]{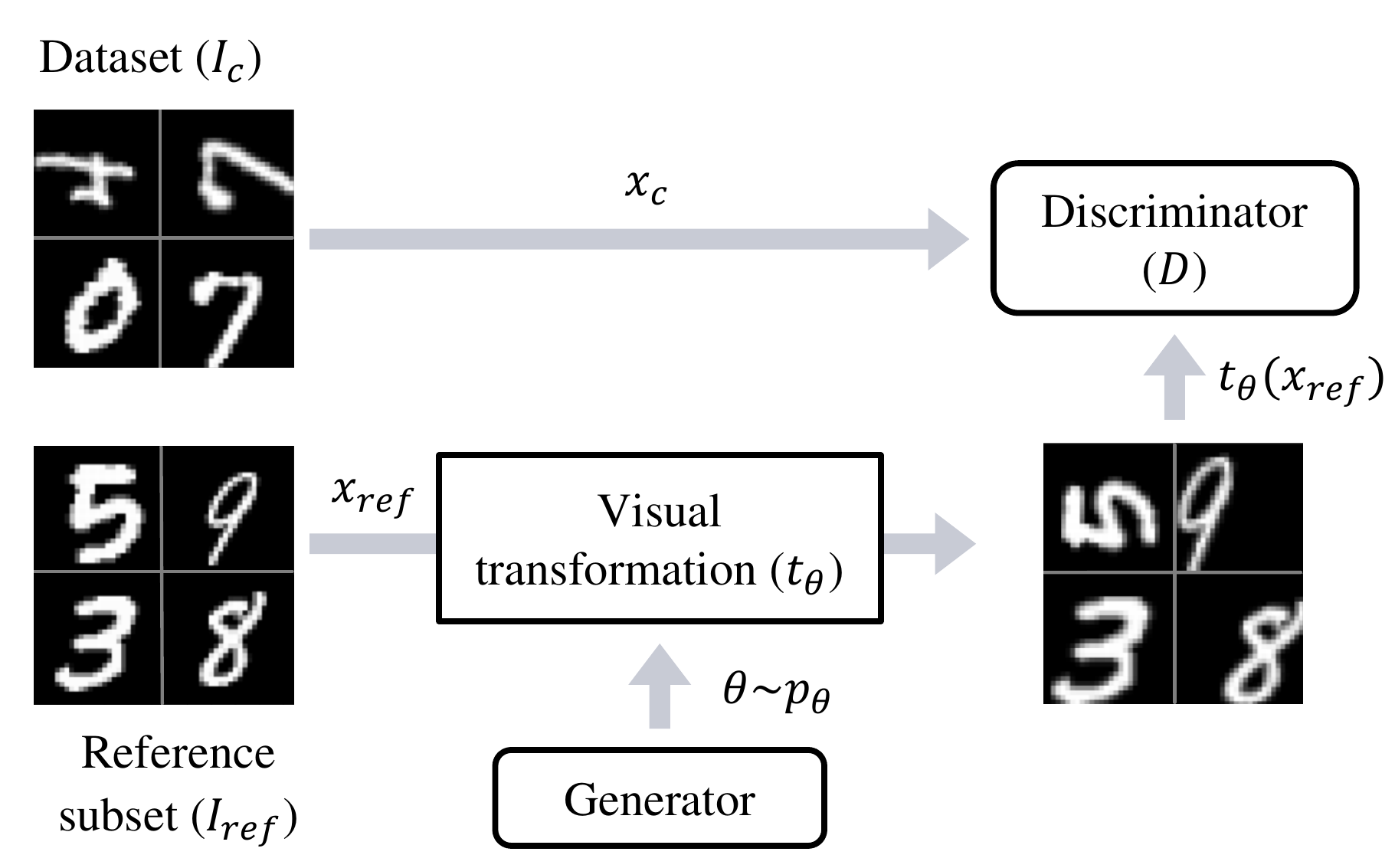}
  \caption{}
  \label{fig1:sfig1}
\end{subfigure}%
\begin{subfigure}{.48\textwidth}
\epsfig{figure=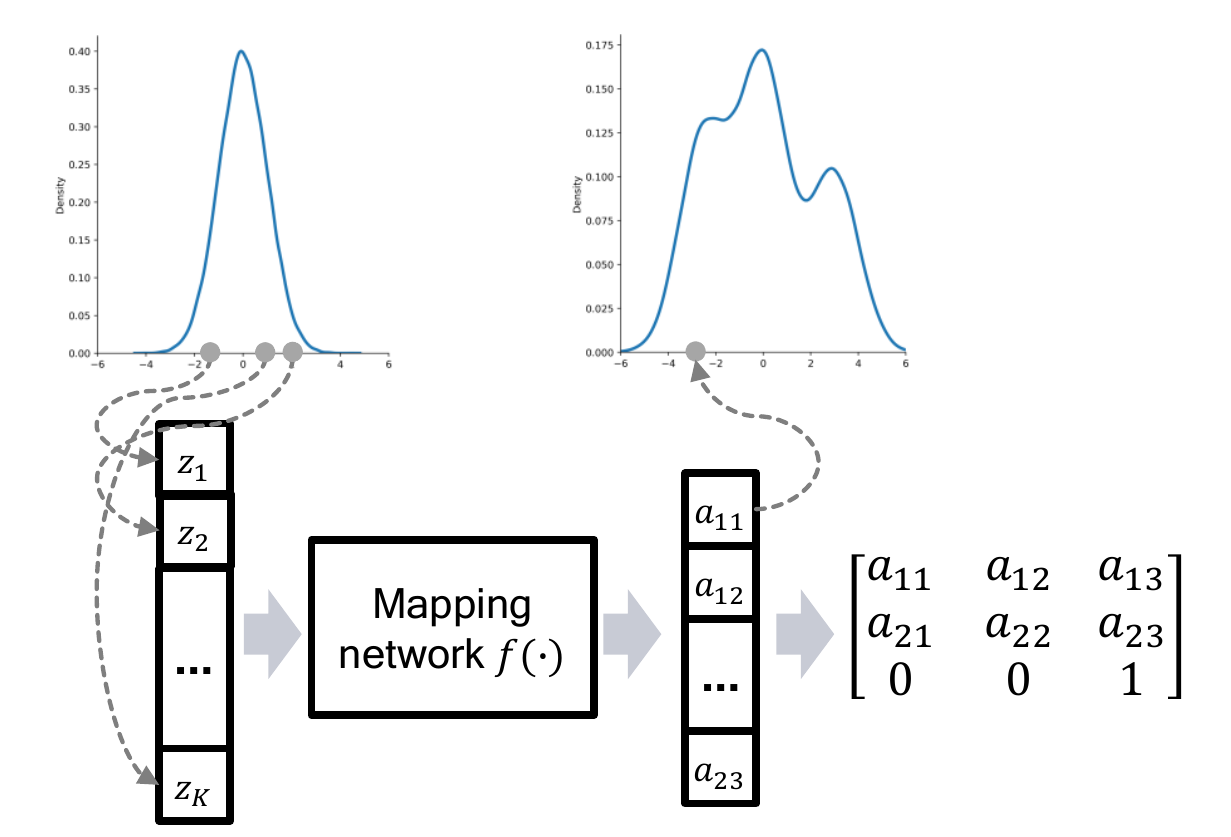,width=1.0\linewidth}
\hspace{-3mm}
  \caption{}
  \label{fig1:sfig2}
\end{subfigure}
\caption{(a) Our framework to estimate the visual transformations present in a dataset $I_c$. (b) Generator produces visual transformation parameters through the mapping network that projects {\it a known} distribution to the desired distribution.
\vspace{-5mm}
}
\label{fig1:fig}
\end{figure}

\section{Proposed Method}
\label{headings}

\subsection{Distribution estimation for visual transformations}
{

}
\paragraph{Learning framework to estimate visual transformations.} 
{We propose a learning-based framework to estimate the distribution of visual transformations present in the dataset. In practice, most collected datasets inherently assume that visual transformations they depict are from specific distributions. Generally, the priors on such distributions are not available, and thus we consider learning a parameterized model to represent such distributions.

 
For a dataset $I_c$, we choose a reference subset that includes the most representative and frequent data ({\it e.g.}, upright images), denoted by $I_{ref}=\{x_{ref}^{i}\}_{i=1}^{N}$ where $N$ is 
much smaller than the size of $I_c$ ({\it e.g.}, 1-3 images per class). Let $\mathcal{T}$ be a set of all possible visual transformations such as affine, color, style-based visual transformation, etc. Assume it consists of $K$ transformations, $\mathcal{T} = \{t_{1}, t_{2}, ..., t_{K}\}$, where $t_{k}$ denotes a transformation type, and $t_{k}(x)$ represents the corresponding transformed data. 
If we augment $I_{ref}$ with the transformations present in $I_c$, it is intuitive that the transformation parameter distributions of $I_{trans}=\{t_{1}(x_{ref}^{1:N}), t_{2}(x_{ref}^{1:N}), ..., t_{K}(x_{ref}^{1:N})\}$ and $I_c$ are similar.
We omit $k$ for the simplicity in describing our algorithm for the subsequent paragraphs.



Based on this, we consider to model $p_{\theta}$ which is the distribution of transformation parameter for $I_{trans}$, from where a specific transformation $t$ can be sampled. Here, we learn $p_\theta$ to represent $p_c$ which is the distribution of transformation parameter for $I_c$, by minimizing the distance between $p_\theta$ and $p_c$ with an adversarial learning (discriminator and generator). The idea is illustrated in Fig.~\ref{fig1:sfig1}.



One of key components of our framework is modeling the learnable distribution $p_{\theta}$ with the parameterized transformation $t(x;\theta)$. A transformation parameter is sampled from $p_{\theta}$ to obtain a specific transformation $t(x;{\theta})$. We consider training a mapping network $f(\cdot)$ to project a standard distribution into the desired distribution $p_\theta$, and the reparameterization trick~\cite{kingma2013auto} is exploited to backpropagate through to the parameters $\theta$ of the distribution $p_{\theta}$. In Appendix~\ref{Visual_T}, we describe how to parameterize visual transformations $t$ including geometric and color transforms 
and how to flow back the loss gradients to sampling grid coordinates for geometric transforms.

As illustrated in Fig~\ref{fig1:sfig2}, we learn the mapping network $f(\cdot)$ that maps a known distribution to the desired distribution. 
This can induce more complex distributions. The mapping network consists of fully connected layers with non-linear activations, and the inputs are random vectors sampled from known distributions, {\it e.g.}, uniform or Gaussian distributions. 
The outputs of the mapping network are transformation parameters, and $p_\theta$ can be obtained from the histogram of the outputs. Thus, the network learns how to produce the desired distributions by combining known distributions.
In summary, the network projects a simple distribution into any complex distribution that can model the distribution of transformations present in the dataset.

To train $p_{\theta}$, we minimize the distance between two distributions $p_{\theta}$ and $p_c$. For this, we consider the following loss function, $L = \underset{\theta \sim p_{\theta} }{E}[l(t({\bf x}_{ref};\theta), {\bf x}_{c})]$,
where $t(;\theta)$ indicates a visual transformation parameterized by $\theta$, and $l$ is a distribution matching objective function such as MMD loss~\cite{gretton2006kernel} and GAN-based loss functions~\cite{goodfellow2014generative, arjovsky2017wasserstein, gulrajani2017improved}. In this work, we employ WGAN-GP~\cite{gulrajani2017improved}, which expedites the optimization of the generator as well as enforce the Lipschitz constraint by using {the} gradient penalty not to fail to converge.

}

\subsection{Automating transformation policies}



\paragraph{A complementary set of current visual transformations.}
We train the network that maps a known distribution to the desired distribution, i.e., the outputs of this network follow $p_\theta$. To approximate this distribution $p_{\theta}$, we use a sampling procedure. Specifically, we feed random vectors sampled from the known distribution to the mapping network to produce a number of output samples, which are then aggregated to form a histogram of values. 
Based on this histogram, its complementary $q_{\theta}$ can be obtained by subtracting each histogram value from the peak histogram value and normalizing it.

\paragraph{Transformation policies for self-supervised tasks.}
We consider two policies for choosing a non-conflicting pretext task: manual and automated policy.
In the manual policy, we sample the transformation instance from the parameter ranges where $p_\theta$ is $0$ or low value.
For example, suppose that $I_c$ contains the rotated images from 0 to 120 degrees. We can sample the transformation instance for the pretext task from 120 to 360 degrees manually.
In the automated policy, the transformation instance is sampled from the constructed complementary distribution $q_\theta$.
To sample a value following a specified distribution, we exploit the {\it inverse transform sampling} method, which is a well-known method for pseudo-random number sampling. First, the CDF of $q_\theta$ is obtained. Then, we obtain the transformation instances by mapping the samples from the uniform distribution $U(0,1)$ to the inverse of the CDF. The obtained transformation instances can then be used for the pretext task.

\begin{figure}
\begin{subfigure}{.35\textwidth}
  \includegraphics[scale=0.45]{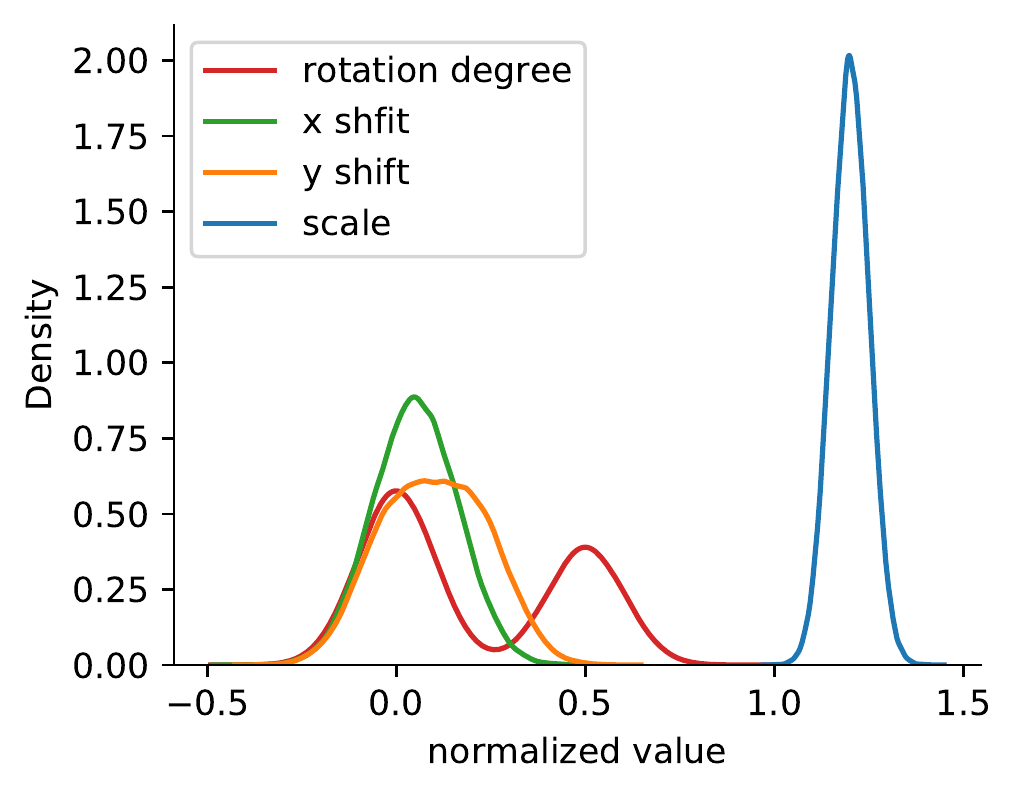}
  \caption{}
  \label{fig:sfig1}
\end{subfigure}%
\begin{subfigure}{.35\textwidth}
  \includegraphics[scale=0.45]{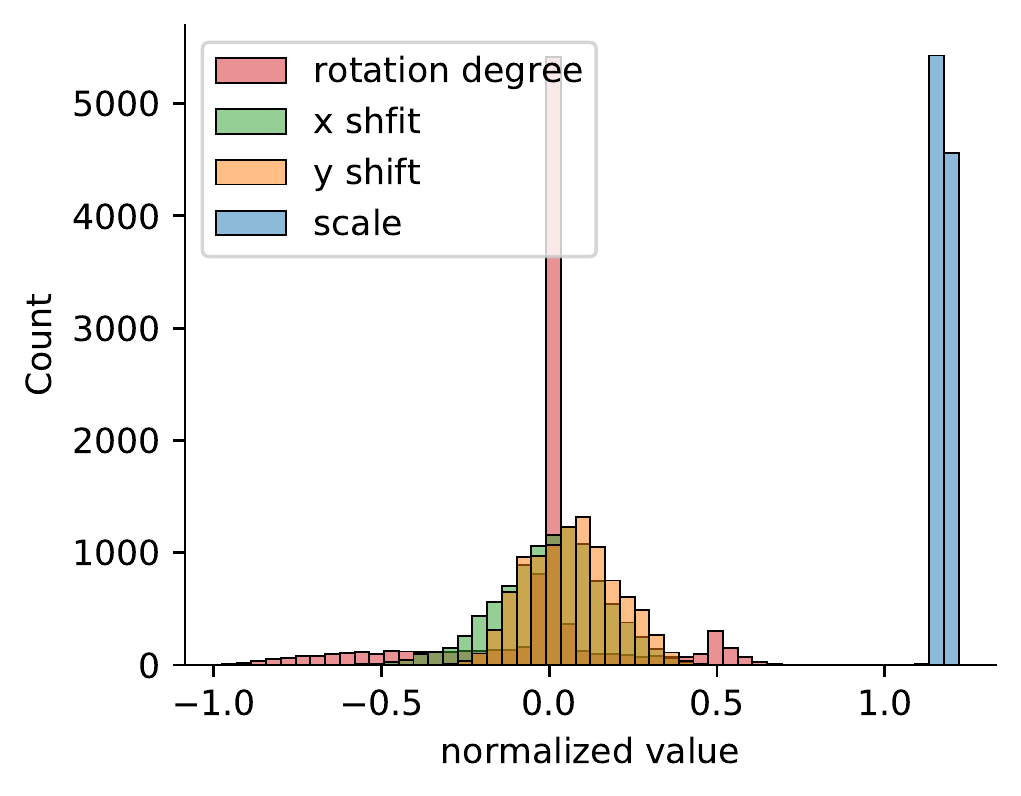}
  \caption{}
  \label{fig:sfig2}
\end{subfigure}%
\begin{subtable}{.3\textwidth}
\vspace{9mm}
\centering
  \resizebox{0.8\textwidth}{!}{
\begin{tabular}{c|c}
\Xhline{2\arrayrulewidth}
Task               & Sup acc. \\ \hline \hline
R: 0, 90           & 92.28    \\ \hline
R: 0, 180          & 96.04    \\ \hline
T: 0.1, 0.1        & 11.35    \\ \hline
T: 0.5, 0.5        & 47.97    \\ \hline
S: x1, x1.25, x1.5 & 33.57    \\ \hline
S: x0.8, x1, x1.5  & 66.57    \\ \Xhline{2\arrayrulewidth}
\end{tabular}}
\vspace{5.2mm}
\caption{}
  \label{fig:sfig3}
\end{subtable}
\caption{(a) The ground-truth distribution of Transformed MNIST. (b) Histogram of output values from the mapping network. (c) Experimental results of the pretext task parameters in the left column. 
Once the representation with the pretext task is learned, the classifier is trained with supervised learning using the fixed representations. We denote this classification accuracy by Sup acc.}
\label{fig:fig}
\end{figure}

\vspace{-5mm}
\section{Experiments}
\vspace{-1.5mm}
We demonstrate that our framework can estimate the distribution of visual transformations present in the dataset. Experimental details are in Appendix~\ref{dis_est_exp}. First, we estimated the distribution on Transformed MNIST, which is constructed by applying various affine transformations to the original MNIST~\cite{726791}.
Fig. \ref{fig:sfig1} illustrates the ground-truth distribution of the transformations present in Transformed MNIST.
Note that the rotation, translation, and scaling operations have their own units of measurements, but we normalize them for better visualization and more stable training: $[-1, 1]$.

For Transformed MNIST, as shown in Fig. \ref{fig:sfig2}, the obtained histogram is similar to the ground truth distribution. Based on this, we can design the pretext tasks for given dataset. In particular, we define two types of self-supervision tasks. The first type is the transformation instance from $p_\theta$, and the second type is that from $q_\theta$, i.e., complementary to $p_\theta$. In Fig. \ref{fig:sfig3}, we compared these two types of pretext tasks where R, T, and S indicate rotation, translation, and scaling, respectively. Once the self-supervised network is trained with pretext tasks, the representations are used for downstream tasks (see Appendix~\ref{self_sup_exp}). For the rotation case, the representations learned with the task classifying 0 and 90 degrees encounter the transformation conflict (we can find 90 degree has high probability from Fig. \ref{fig:sfig1}) and result in performance degradation compared to the task of 0 and 180 degree. Especially, in the case of translation and scaling, the performance gap becomes bigger.
The results demonstrate that complementary pretext tasks lead to learn useful representation for downstream task.

\begin{wrapfigure}{r}{6.0cm}
\vskip -0.18in
\begin{subfigure}{\textwidth}
\centering
  \includegraphics[scale=0.41]{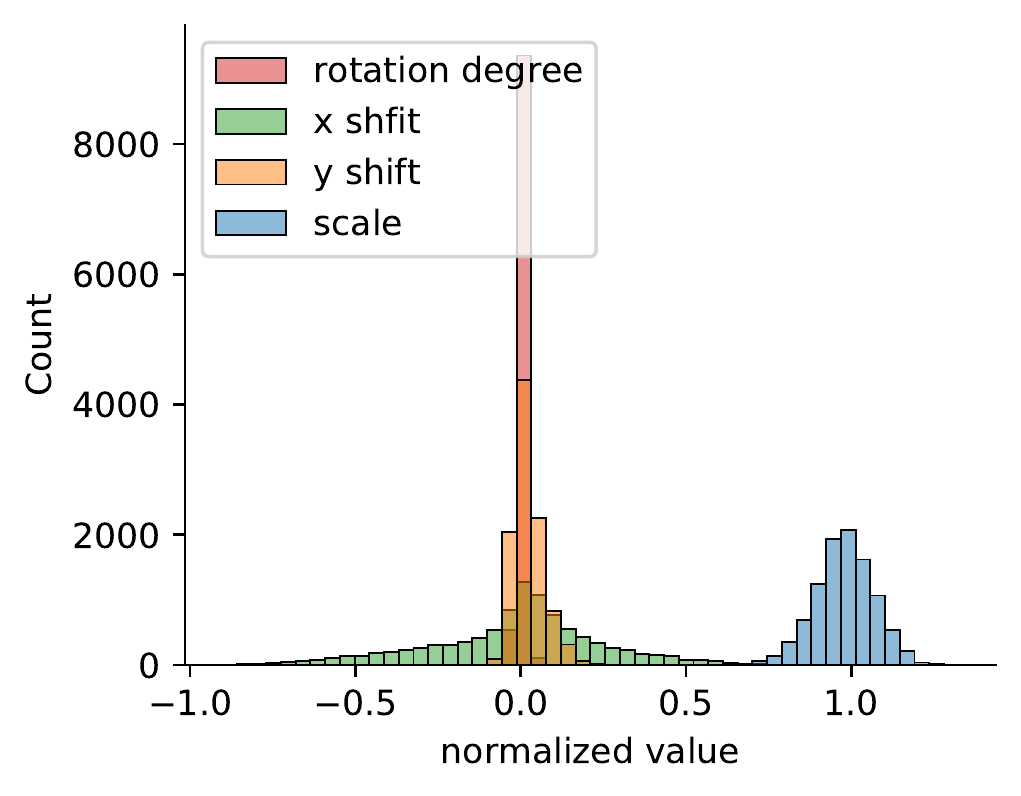}
\end{subfigure}
\begin{subtable}{\textwidth}
\centering
  \resizebox{0.5\textwidth}{!}{
\begin{tabular}{c|c}
\Xhline{2\arrayrulewidth}
Task               & Sup acc. \\ \hline \hline
S: x1, x1.14, x1.33      & 54.30    \\ \hline
S: x1, x1.23, x1.6        & 64.95    \\ \hline
S: x1, x1.5, x2  & 67.03    \\ \hline
T: 0.1, 0.1 & 92.34    \\ \hline
T: 0.5, 0.5 & 93.63    \\ \Xhline{2\arrayrulewidth}
\end{tabular}}
\end{subtable}
\caption{Experimental results on SVHN.}
\label{fig5}
\end{wrapfigure}

Next, we applied our automated policy to the real dataset. We plot the histogram of transformation parameters for the SVHN dataset in Fig.~\ref{fig5}.
In VTSS~\cite{pal2019towards}, they described the results of pretext tasks by using prior knowledge or manual inspection on the dataset. 
In our case, we analyze the results based on the estimated distribution beyond the prior knowledge available for the dataset and define effective pretext tasks even significant degree of transformations already exist in the dataset.
In VTSS, they observed that the pretext task based on scale-based transformations leads poor performance for SVHN.
This is because the SVHN dataset has some degrees of scale variation (refer the histogram in Fig.~\ref{fig5}).
More experimental results using other datasets are described in Appendix~\ref{more_exp}.

\vspace{-2mm}
\section{Conclusion}
\vspace{-2mm}
We propose a framework to automatically estimate the distribution of visual transformations present in the dataset. This enables efficiently creating pretext tasks that are not depicted in the dataset. We show that the representations learned from this task are more informative on downstream classification scenarios compared with those from the conflicting task.

\bibliographystyle{unsrt}
\bibliography{neurips_2021_workshop.bib}

\appendix

\section{Distribution estimation for visual transformations}
\label{dis_est_exp}
\subsection{Implementation details}
Our framework to estimate \sr{the} distribution of visual transformations consists of two networks: a generator and a discriminator. The generator 
\sr{is constructed with} three linear layers with 128 dimensional output 
\sr{except that the last layer is 6 dimensional output.} We used LeakyReLU with $0.2$ slope for the first two layers, and the last layer has the tangent hyperbolic function to produce values 
\sr{between $-1$ and $1$.}
The generator takes 10 dimensional inputs sampled from Gaussian distribution, and outputs 6 parameters for affine transformation \sr{(rotation, scale, and translation). We found that the affine transformation parameters can be accurately estimated with two generators: {\it i.e.}, the first one for the scale parameter and the other one for remaining five parameters. }
Similar with the affine case, \sr{the color transformation can be generated from 10 dimensional inputs (See the details in Appendix~\ref{Visual_T}). In this case, the generator outputs four dimensional parameters: brightness, saturation, contrast, and hue. And, the network architecture is the same with the affine transformation case, but the output dimension is four.} 
We adopted a simple discriminator that has three linear layers with the LeakyReLU function.
\sr{The input size of the first layer depends on the image size, and the output dimensions of the three layers are respectively 512, 256, and 1.} The \sr{discriminator} output 
represents a fake or a real, {\it i.e.}, an image from the \sr {given} dataset \sr{(real)} or a transformed image using the generator \sr{(fake)}.

\begin{figure}[!htbp]
\begin{subfigure}{.33\textwidth}
\centering
  \includegraphics[scale=0.4]{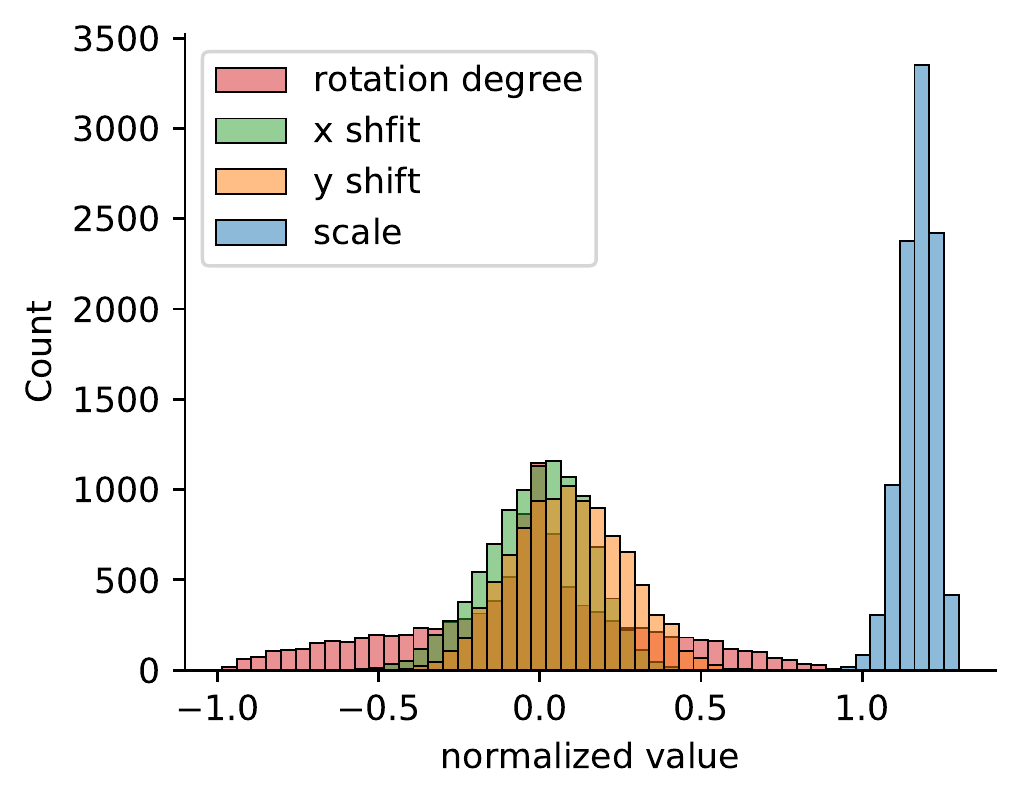}
  \caption{At 100,000 iterations}
\end{subfigure}%
\begin{subfigure}{.33\textwidth}
\centering
  \includegraphics[scale=0.4]{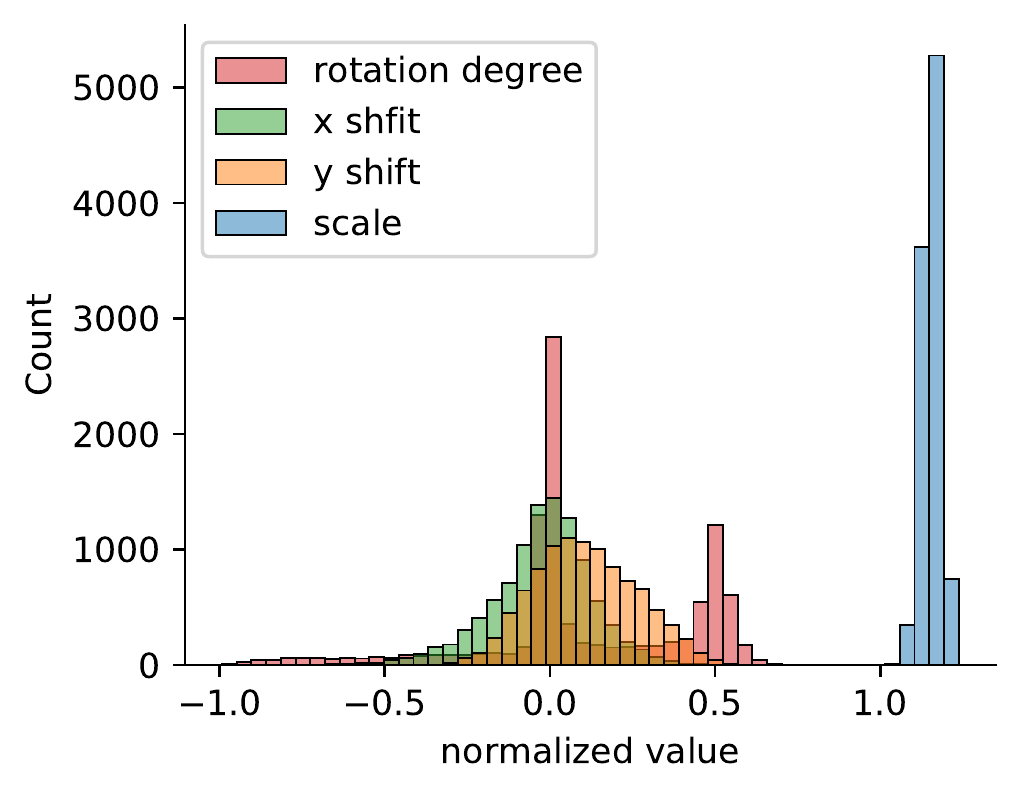}
  \caption{At 500,000 iterations}
\end{subfigure}%
\begin{subfigure}{.33\textwidth}
\centering
  \includegraphics[scale=0.4]{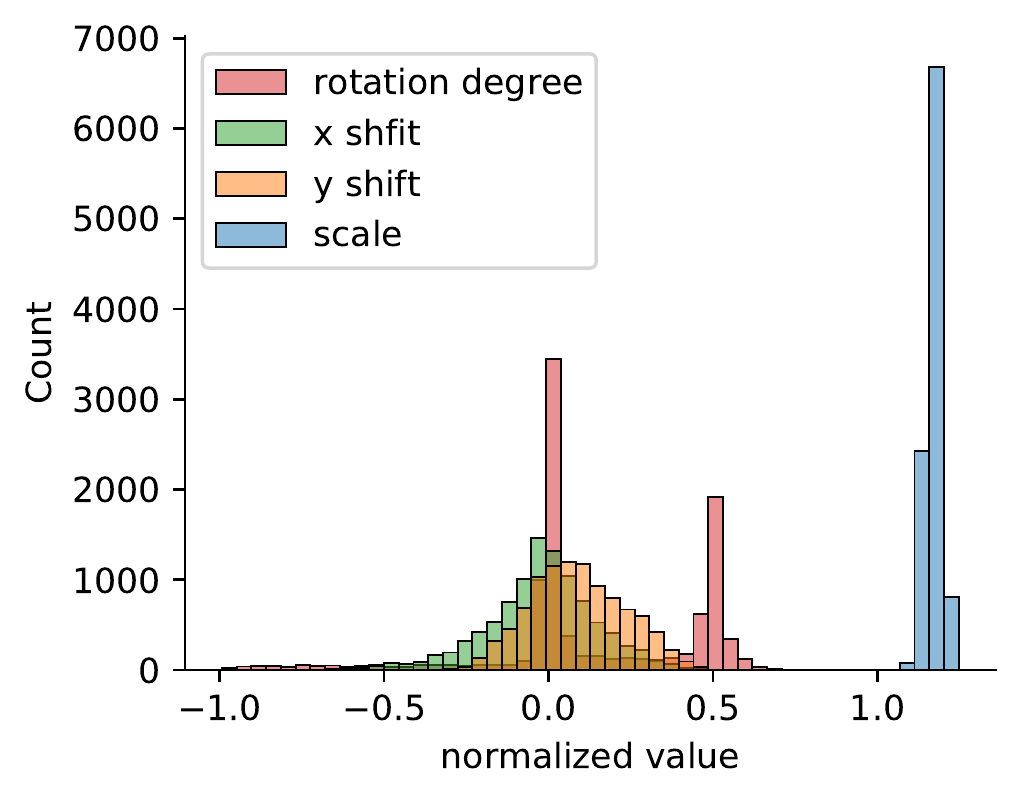}
  \caption{At 1,000,000 iterations}
\end{subfigure}%
\caption{
\sr{The histograms obtained from the outputs of}
the mapping network \sr{trained using} Transformed MNIST at (a) 100,000, (b) 500,000, and (c) 1,000,000 iterations.}
 \label{fig22}
\end{figure}

We trained these networks with the aid of WGAN-GP~\cite{gulrajani2017improved}.
We adopted the basic settings \sr{of} WGAN-GP: $\lambda=10$, $n_{critic}=5$, $\alpha=0.0001$, $\beta_{1} = 0$, and $\beta_{2} = 0.9$. \sr{With the batch size of $10$ and learning rate of $5e-5$, we} 
trained the network 
for $500,000$ iterations, which \sr{took} $6$ hours \sr{in our single} NVIDIA TITAN Xp GPU.
\sr{Fig.~\ref{fig22} illustrates three histograms obtained at different iteration points, and we observed that there is  no big difference after 500,000 iterations.}

\subsection{The method to reduce the artifact of a transformed image} \label{artifact_generator}
Through adversarial learning, the discriminator is trained to distinguish the original images in the dataset \sr{from the images transformed by the generator given the reference subset.}
Our objective is to train the generator to deceive the discriminator such that the transformed images have a similar distribution with the original images.
However, 
\sr{the discriminator may easily identify the generated images when they have} artifacts like zero padding. \sr{These artifacts cause the network to learn an undesired property.}
For example, assume that the input image contains a dog \sr{of} 45 degree rotated, 
\sr{and the objective is to find the amount of the rotation by generating a similar image using an upright dog image. However, when the image is rotated in the generator,} 
the four corners are \sr{zero-padded, and hence it} provides a crucial hint for the discriminator. \sr{Often, it leads to converge to a meaningless distribution which is not our objective.}
To mitigate this problem, we perform transformations first \sr{which is followed by a center-crop.}
\sr{For example, given a 32x32 image,} 
\sr{a transformation is applied, and then the 24x24 image is obtained by the center-crop}, which can cover 8 pixel shifts with no zero padding. For 256x256 image, \sr{the resulting image size is 196x196} 
where these two numbers \sr{were} empirically selected through multiple trials. \sr{With this procedure, we can accurately estimate}
the distribution of visual transformations \sr{without an artifact effect.} 

\section{Self-supervised learning for downstream tasks}
\label{self_sup_exp}
\subsection{Implementation details}
We used the NIN architecture~\cite{lin2013network} as the backbone feature extraction network to learn \sr{the} representations for downstream tasks.
The NIN consists of four convolutional blocks, and each block contains three convolutional layers. 
Convolution, batch normalization, and ReLU are sequentially placed in a single convolutional layer.
\sr{In training a pretext task for rotation, translation, and scaling categories, the classifier of a single linear layer is added after global average pooling.} 
We used SGD with the batch size of $128$, momentum of $0.9$, weight decay of $5e-4$, and learning rate of $0.1$.
We dropped the learning rates by a factor of $5$ after epochs $60$, $120$, and $160$.
We trained the network in total for $200$ epochs.
For a fair comparison among several pretext tasks, we used the same training procedure.

\sr{To evaluate the representation learned by self-supervision, a classifier is trained using the learned features.} 
Specifically, in our experiments on Fashion MNIST~\cite{xiao2017/online}, SVHN~\cite{netzer2011reading}, CIFAR-10~\cite{krizhevsky2009learning}, and CIFAR-100~\cite{krizhevsky2009learning}, we followed the existing evaluation protocols~\cite{gidaris2018unsupervised, pal2019towards} by adding a classifier on top of the second convolutional block.
The classifier consists of a single conv block, \sr{global average pooling, and one linear layer. Here, the conv block is the same with the conv3 in NIN architecture.}
We used SGD with batch size $128$, momentum $0.9$, weight decay $5e-4$, and learning rate of $0.1$, which is dropped by a factor of $5$ after epochs $35$, $70$, and $85$.
We trained the network in total for $100$ epochs.

\subsection{Details of constructing pretext tasks}
We introduce four pretext tasks, R, T, S, and RST, which indicate rotation, translation, scaling, and the combination of all, respectively. For the rotation case, the unit is the rotation degree. Translation and scaling units are the ratio of shift amount to the image size and the scaling ratio.
Note that the translation task involves 5-way classification: left and right shift along with $x$ and $y$ axis, and 0 translation.

When constructing a pretext task of rotation, translation, and scaling, we also encounter the same artifact problem. The artifact induced by the transformation for a pretext task would be a big hint for the network, and hence the useful representation for downstream tasks may not be obtained.
\sr{With this reason, we consider the pretext tasks of each transformation as follows: 1) For rotation, we use 90, 180, 270, and 360 degree which do not incur the zero-padding, 2) For translation and scaling, we perform the center crop depending on the task parameter: e.g. when we define the translation task of 3 pixels in 32x32 image, we crop the center image of 26x26. }

\section{Visual transformation}
\label{Visual_T}
\subsection{Parameterize geometric transformation}
We describe how to parameterize a distribution over the set of visual transformations, particularly affine transformations. This discussion can be extended to projective and spline transformations as well. For the affine transformation, we include translation, scaling, 
similarity, reflection, rotation, and shear. The affine matrix is parameterized as follows:
\begin{equation}
t =
\begin{bmatrix}
a_{11} & a_{12} & a_{13}\\ 
a_{21} & a_{22} & a_{23}\\ 
0 & 0 & 1
\end{bmatrix}, \quad \text{where} \quad a_{11},a_{12}, ...,a_{23} \in \theta.
\label{affine}
\end{equation}
These parameters are sampled from $p_{\theta}$, and this matrix is applied to the coordinate system. We adapt Spatial Transformer Network~\cite{jaderberg2015spatial} allowing loss gradients to flow back to the sampling grid coordinates as well as to the input image. For example, in case of affine transformation, the sampling grid is result of warping the regular grid with an affine matrix, as in~(\ref{affine}). If we use a bilinear sampling kernel to map the output pixel value, the output image $V$ is obtained as follows: 
\begin{equation}
O_{i}^{c} = \sum_{n}^{H} \sum_{m}^{W}I_{nm}^{c} \text{max}(0, 1-\left | u_{i}^{s}-m \right |) \text{max}(0, 1-\left | v_{i}^{s}-n \right |),
\end{equation}
where $O_{i}^{c}$ indicates the output value for pixel $i$ at location $(u_{i}^{t}, v_{i}^{t})$ of channel $c$, and $(u_{i}^{t}, v_{i}^{t})$ are the target coordinates of the regular grid. $I_{nm}^{c}$ is the value at location $(n, m)$ of the input image with height $H$ and width $W$, and $(u_{i}^{s}, v_{i}^{s})$ indicates the spatial location of the input corresponding to the output coordinate through the sampling grid. We can define the gradients with respect to $I_{nm}^{c}$ and $(u_{i}^{s}, v_{i}^{s})$ such that the backpropagation of the loss can be flowed through this sampling mechanism.


\subsection{Parameterize color transformation}
\sr{In all previous experiments, we considered the transformations of rotation, scale, and translation. This section will describe how we can parameterize the color transformation. }
Specifically, we consider \sr{changing the} 
brightness, saturation, contrast, and hue of an image.
We introduce learnable parameters for these transformations and train the mapping network \sr{such that each parameter represents} 
the transformation distribution of the dataset.

\sr{First}, the brightness 
\sr{change can be simply expressed as follows:}
\begin{equation}
x_{brt} = x\alpha_{brt},
\end{equation}
where $\alpha_{brt}$ is a learnable scale factor.
\sr{Second, we can change} the saturation of $x$ by using a linear combination of the original image $x$ and the gray-scaled version of the image $x_{gray}$ as follows:
\begin{equation}
    x_{sat} = x\alpha_{sat} + x_{gray}(1-\alpha_{sat}), \quad \text{\sr{and}} \quad x_{gray}= 0.299x_{r} + 0.587x_{g} + 0.114x_{b},
\end{equation}
where $x_{r}$, $x_{g}$, and $x_{b}$ indicate three color channels of $x$. \sr{The scale factor} $\alpha_{sat}$ is in the range of $[0, 1]$. \sr{And third,} we can \sr{change the} contrast of $x$ by calculating a linear combination of $x$ and the average of $x_{gray}$ over all spatial dimensions as follows:
\begin{equation}
    x_{con} = x\alpha_{con} + mean(x_{gray})(1-\alpha_{con}),
\end{equation}
where $\alpha_{con}$ is a scale parameter.
Finally, we resort to a linear approximation for the hue \sr{change} in the YIQ color space.
We firstly convert RGB to YIQ, and then apply rotation to the IQ components.
The transformation from RGB to YIQ is denoted by $T_{YIQ}$, and 
\sr{we can obtain the following expression}:
\begin{equation}
    \begin{bmatrix}
Y\\ 
I\\ 
Q
\end{bmatrix} =
T_{YIQ}
\begin{bmatrix}
R\\ 
G\\ 
B
\end{bmatrix}, \quad \text{where} \quad T_{YIQ} = 
\begin{bmatrix}
 0.299& 0.587 & 0.114\\ 
 0.596& -0.275 & -0.321\\ 
 0.212& -0.523 & 0.311
\end{bmatrix}.
\end{equation}
In the YIQ color space, we \sr{can change} the hue of an image by rotating the IQ components \sr{with} 
a rotation matrix:
\begin{equation}
R_{\theta_{hue}}=
\begin{bmatrix}
 1& 0 & 0\\ 
 0& \text{cos}\theta_{hue} & -\text{sin}\theta_{hue}\\ 
 0& \text{sin}\theta_{hue} & \text{cos}\theta_{hue}
\end{bmatrix},
\end{equation}
where $\theta_{hue} = 2\pi \alpha_{hue}$, and $\alpha_{hue}$ is a learnable parameter.
In summary, the hue \sr{change can be expressed} 
by the following matrix multiplication:
\begin{equation}
x_{hue} = T_{RGB}R_{\theta_{hue}}T_{YIQ}x,
\end{equation}
where $T_{RGB}$ is the inverse matrix of $T_{YIQ}$.
Using the procedure, we can 
\sr{employ} learnable parameters for color transformation, and hence we can estimate the distribution of color transformations present in the dataset.

\section{Experimental results}
\label{more_exp}
\subsection{Distribution estimation}

\begin{wrapfigure}{r}{7.0cm}
\vskip -0.24in
\begin{subfigure}{\textwidth}
\centering
  \includegraphics[scale=0.41]{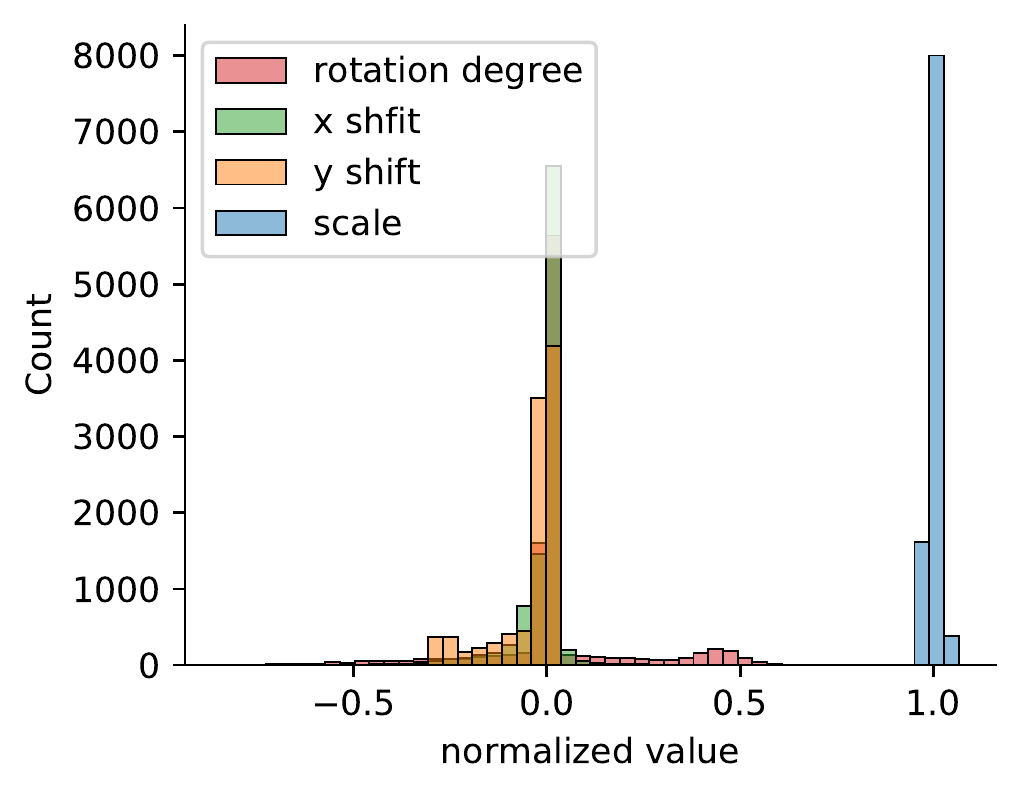}
\end{subfigure}
\caption{Experimental results on Fashion MNIST.}
\vskip -0.18in
\label{fig6}
\end{wrapfigure}

We plot the histogram of transformation parameters for Fashion MNIST~\cite{xiao2017/online} in Fig.~\ref{fig6}.
As illustrated in Fig.~\ref{fig6}, the Fashion MNIST dataset consists of well-aligned images, i.e., almost no transformation is present. This suggests that the network can learn powerful representations with pretext tasks constructed by any visual transformation.

Furthermore, we illustrated the histograms of output values obtained from the mapping network on four visual recognition datasets, CropDisease~\cite{mohanty2016using}, EuroSAT~\cite{helber2019eurosat}, ISIC2018~\cite{tschandl2018ham10000, codella2019skin}, and ChestX~\cite{wang2017chestx}.

\begin{figure*}
    \centering
    \begin{subfigure}[b]{0.475\textwidth}
        \centering
        \includegraphics[width=0.7\textwidth]{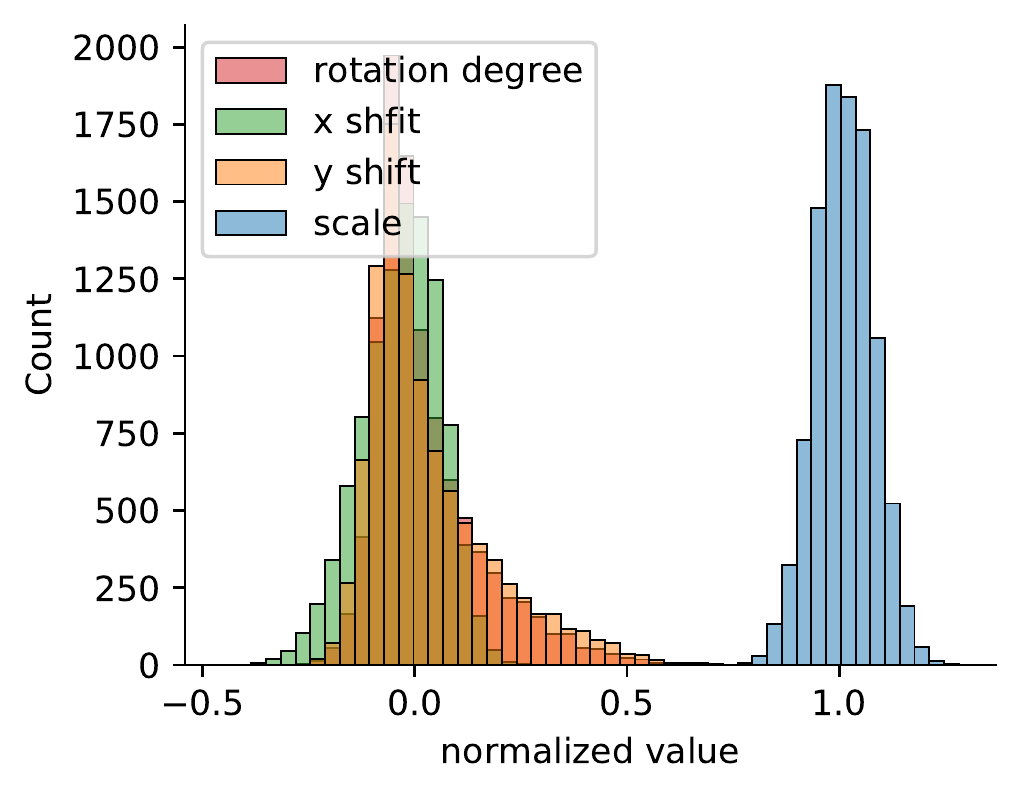}
        \caption[]%
        {{\small CropDisease}}    
        \label{fig:mean and std of net14}
    \end{subfigure}
    \hfill
    \begin{subfigure}[b]{0.475\textwidth}  
        \centering 
        \includegraphics[width=0.7\textwidth]{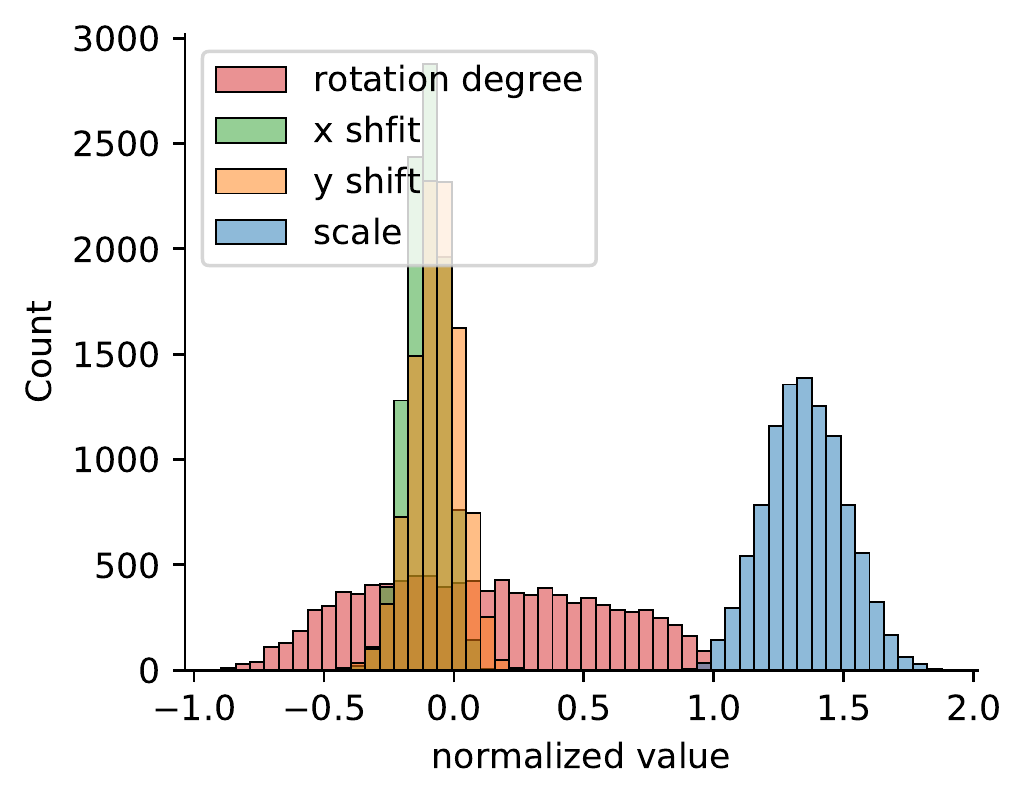}
        \caption[]%
        {{\small EuroSAT}}    
        \label{fig:mean and std of net24}
    \end{subfigure}
    \vskip\baselineskip
    \begin{subfigure}[b]{0.475\textwidth}   
        \centering 
        \includegraphics[width=0.7\textwidth]{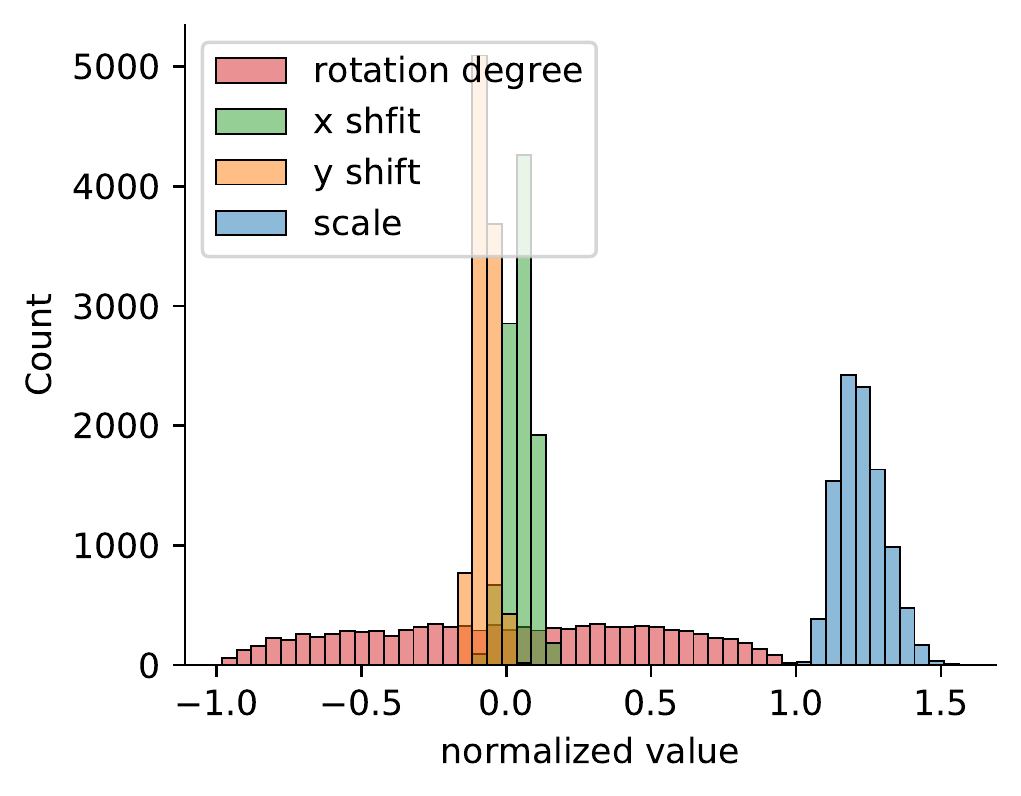}
        \caption[]%
        {{\small ISIC2018}}    
        \label{fig:mean and std of net34}
    \end{subfigure}
    \hfill
    \begin{subfigure}[b]{0.475\textwidth}   
        \centering 
        \includegraphics[width=0.7\textwidth]{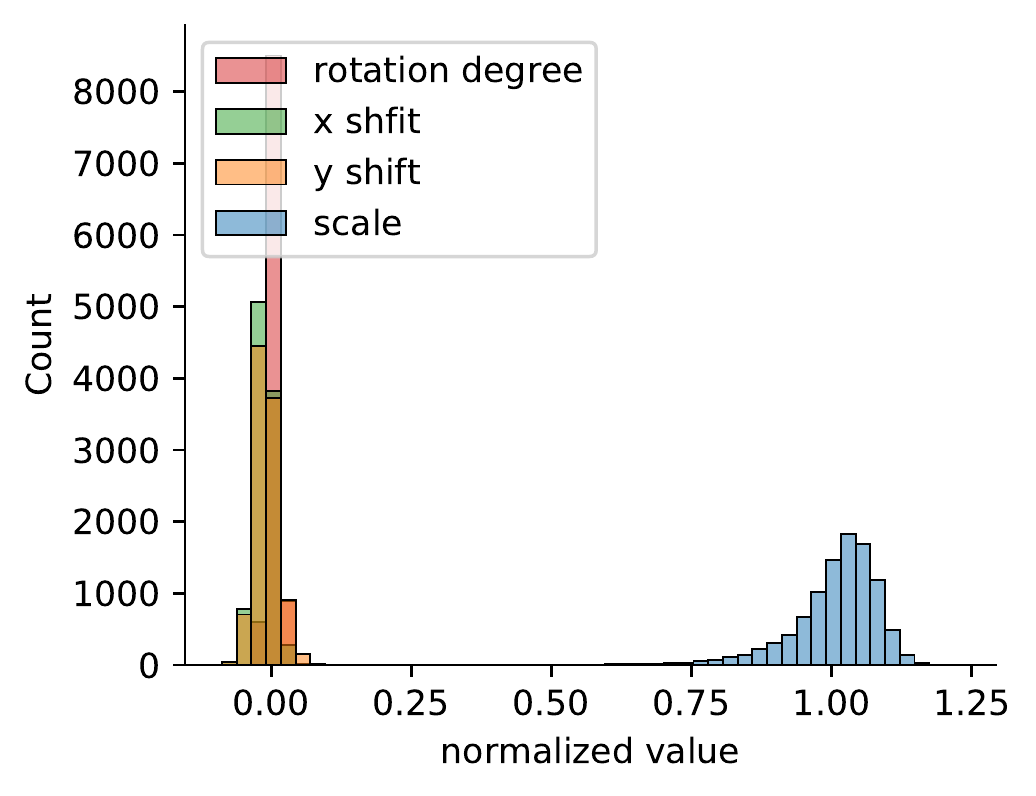}
        \caption[]%
        {{\small ChestX}}    
        \label{fig:mean and std of net44}
    \end{subfigure}
    \caption[ The average and standard deviation of critical parameters ]
    {The \sr{estimated} distributions of visual transformation on CD-FSL benchmarks.} 
    \label{fig:CDFSL}
\end{figure*}

Even when the images are big, we found that the mapping network can represent the transformation distribution well as shown in Fig~\ref{fig:CDFSL}.
The EuroSAT and ISIC2018 dataset have the rotation transformation almost over the entire range, $[-\pi, \pi]$. Visually checking the image samples, we can find the two datasets already contain various rotated images.

\subsection{Comparison with other methods on downstream tasks}

\begin{wraptable}{r}{7.0cm}
\caption{Performance comparison on downstream tasks.}
\begin{subtable}{\textwidth}
\centering
\resizebox{0.85\textwidth}{!}{
\begin{tabular}{c|c|c|c|c}
\Xhline{2\arrayrulewidth}
         & FMNIST & SVHN  & CIFAR-10 & CIFAR-100 \\ \hline \hline
RotNet & 91.94  & 91.29 & 89.15    & 63.62     \\ \hline
VTSS   & 92.18  & 91.72 & 89.58    & 64.87     \\ \hline
Ours   & 93.22  & 94.75 & 89.58    & 64.00     \\ \Xhline{2\arrayrulewidth}
\end{tabular}}
\caption{Comparison with other methods.}
\label{comp_with_others}
\end{subtable}
\begin{subtable}{\textwidth}
\centering
\resizebox{0.85\textwidth}{!}{
\begin{tabular}{c|c|c|c|c}
\Xhline{2\arrayrulewidth}
         & FMNIST & SVHN  & CIFAR-10 & CIFAR-100 \\ \hline \hline
Baseline & 91.63  & 91.72 & 86.32    & 60.79     \\ \hline
Ours   & 93.22  & 94.75 & 87.01    & 62.48     \\ \Xhline{2\arrayrulewidth}
\end{tabular}}
\caption{Comparison with our baselines.}
\label{comp_with_baseline}
\end{subtable}
\end{wraptable}

To compare with RotNet~\cite{gidaris2018unsupervised} and VTSS~\cite{pal2019towards}, we set hyper-parameters with the same condition as others (using horizontal flip for data augmentation). We reported the accuracy in Table 1a.

We want to show that the pretext tasks selected by our estimated distribution can force the network to learn more meaningful representations than the manually selected one.
In Table 1b,
we define the pretext task, RST, which is the combined pretext task of three types of visual transformations.
Baseline selects RST manually and ours defines RST based on the estimated distribution.
We use rotation with 0, 90, 180, and 270, translation with 0.1, 0.1, and scaling x1, x1.14, x1.33 for the baseline RST while we define the pretext task carefully based on estimated distribution. Note that we do not use data augmentation technique to show the effectiveness of each pretext task separately.

\section{Discussion}
\subsection{Extension to contrastive loss}
We consider a contrastive learning framework based on SimCLR~\cite{chen2020simple}. We observed that the contrastive loss with the augmented versions appropriate to the target dataset helps the network to learn generalizable representations even when the domain gap between source and target is large, and target data is scarce. Intuitively, this indicates proper pretext tasks for contrastive loss functions are effective mechanisms to derive meaningful representations for downstream tasks.

We apply the proposed automated transformation policy to generating augmented versions used for contrastive learning. As such, we retrain the base network, ResNet-50~\cite{he2016deep}, with the standard supervised loss on ImageNet~\cite{deng2009imagenet}, then fine-tune the base network, the added classifier, and the projector on the target dataset with both cross-entropy and contrastive losses.
Specifically, we use supervised contrastive loss function~\cite{khosla2020supervised} with the SimCLR~\cite{chen2020simple} framework.
It is to be noted that supervised contrastive loss function builds on the contrastive self-supervised literature by leveraging label information.
We conduct experiments on cross-domain benchmark datasets (CropDisease~\cite{mohanty2016using}, EuroSAT~\cite{helber2019eurosat}, ISIC2018~\cite{tschandl2018ham10000, codella2019skin}, and ChestX~\cite{wang2017chestx}) where we split the total dataset such that $5\%$ was used for training while the remaining was used for testing.

We compare four methods: fine-tuning the network using (a) cross-entropy loss (Baseline) and with additional contrastive loss of (b) random rotation from -180 to 180 (SimCLR with Rot.), (c) random affine transformations, which contain rotation from -180 to 180, translation from -0.5 to 0.5, and scaling from 0.5 to 1.5 (SimCLR with Aff.) and (d) automated transformation policies within the same range of (c) and obtained from the target dataset (SimCLR with ATP). Table \ref{tab:second_table} shows the results of fine-tuning with the contrastive loss function. We found that some pretext tasks, {\it i.e.}, data augmentations, are not helpful to the target tasks as pointed out by~\cite{xiao2021what}. Our automated transformation policies for contrastive learning leads to better performance than the naive approach of using random augmentation. It shows that not only types of transformations, {\it e.g.}, color or spatial transformation, but also the degree of transformations, {\it e.g.}, 30 or 60 degrees rotation impact the performance significantly in contrastive learning.

\begin{table}[!htb]
\caption{Effect of fine-tuning on cross-domain benchmark datasets. Rot., Aff., and ATP indicate random rotation, random affine transformation, and automated transformation policies, respectively.}
\label{tab:second_table}
\resizebox{0.75\columnwidth}{!}{
\begin{tabular}{c|c|c|c|c|c}
\Xhline{2\arrayrulewidth}
Method                  & ChestX & ISIC & EuroSAT & CropDiseases \\ \hline \hline
Baseline                & 41.16     & 67.00   & 96.35& 96.43  \\ \hline
+ SimCLR with Rot.      & 38.24     & 68.43  & 97.57 & 97.48  \\ \hline
+ SimCLR with Aff.      & 40.65     &  68.97   & 97.66 & 97.41  \\ \hline
+ SimCLR with ATP       & 42.33   &  72.97  & 97.72 & 97.60 \\ \Xhline{2\arrayrulewidth}
\end{tabular}}
\end{table}

\subsection{Large-scale experiment}
In the future, we will conduct experiments with pretraining on a large-scale dataset and transfer to other datasets for evaluation. Since the large-scale dataset consists of images following a complex distribution, the manual selection of pretext tasks cannot be the best one and it may produce transformation conflict. Still, we believe that our policies will work well on a large-scale dataset.

\end{document}